# Rapid grasping of fabric using bionic soft grippers with elastic instability*

Zechen Xiong [1, *], Zihan Guo[2], Li Yuan[2], Yufeng Su[2], Yitong Liu[2], Hod Lipson[2]

*Abstract*— Robot grasping is subject to an inherent tradeoff: Grippers with a large span typically take a longer time to close, and fast grippers usually cover a small span. However, many practical applications of grippers require the ability to close a large distance rapidly. For example, grasping cloth typically requires pressing a wide span of fabric into a graspable cusp. Besides, the ability to perform human-like grasping and ease offabrication are also very important for new soft grippers. Here, we demonstrate a human-finger-inspired snapping gripper that exploits elastic instability to achieve rapid and reversible closing over a wide span. Using prestressed semi-rigid material as the skeleton, the gripper fingers can widely open (86 mm) and rapidly close (46 ms) following a trajectory similar to that of a thumb-index finger pinching, and is 2.7 times and 10.9 times better than the reference gripper in terms of span and speed, respectively. We theoretically give the design principle, simulatively verify the method, and experimentally test this gripper on a variety of rigid, flexible, and limp objects and achieve good mechanical performance.

## I. INTRODUCTION

Human hands have approximately 27 degrees of freedom (DOF) and a large number of muscles [1], which enables them to manipulate all kinds of objects with strength, dexterity, and delicacy. Most robots grasp through a complex system of motors which makes the resulting designs inevitably rigid and bulky [2]. Soft robotic grippers, also known as compliant or adaptive grippers, on the other hand, are a new type of robotic end-effector mimicking hands and animal tentacles for handling delicate or fragile objects [3]. Due to the flexible materials used, these grippers can conform to the shape of the object being grasped and apply gentle forces without causing damage. The advantage of being bio-compatible, human-friendly, and highly adaptive gives them an important role in next-generation robotics. Becker et al. use entanglement grasping to circumvent challenges of object recognition, grasp planning, and feedback, via filament-like soft actuators [4]. Yang et al. present a soft gripper using a simple material system based on kirigami shells [5]. Brown et al. design a universal robotic gripper based on the reversible jamming effect of granular material [6]. However, the moduli of the materials used for these grippers usually range from 10-1000 kPa [7], which is smaller than a hundred thousandth the modulus of steel used for rigid robots. This intrinsic softness of materials as well as the actuation methods of soft robotics makes their grasp weak in force and slow in motion, which is also far from the strength and dexterity of human hands.

Bi-stable grippers are systems that store elastic energy and release it in a short amount of time, achieving mechanic performance such as large force and high speed. Besides, bi-stability also helps these systems to save energy since no energy supply is needed to maintain their grasping. Due to these features, bi-stable grippers are thriving in recent years. Zhang et al. design a soft gripper with carbon-fiber-reinforced polymer laminates that snaps within 112 ms [8]. Thuruthel et al. build a soft robotic gripper with tunable bistable properties for sensor-less dynamic grasping which capture objects within 0.02 s [9]. Wang et al. present a soft gripper based on a bi-stable dielectric elastomer actuator (DEA) that only takes 0.17 s and 0.1386 J to snap [10]. Zhang et al. present a 3D-printed bi-stable gripper that allows a palm-size quadcopter to perch on cylindrical objects. [11]. McWilliams et al. develop a push-on-push-off bi-stable gripper that reduces the complexity of grasping [12]. Estrada et al. and Hawkes et al. present a bi-stable gripper with gecko-inspired adhesives to grasp spinning objects [13], [14]. Lerner et al. propose a bi-stable gripper with a variable stiffness [15]. Most of these bi-stable grippers have complex fabrication procedures and limited adaptivity. For example, none of them can emulate the ability of human hands to pick up a piece of fabric. To address these challenges, we propose to use the prestressed bi-stable hair-clip mechanisms (HCM) [16], [17] as end effectors for robotic manipulation. In [16], we derive the theory to calculate the deformation, energy profile, and dynamics of HCMs, and verify it with two designs of robotic fish; in [17], we make adaptation of the theory and apply it on a soft robotic crawler. Here, we are adapting the HCM method to soft manipulators instead of robots.

Compared with other bi-stable grippers, the simple structure, easy fabrication, and convenient assembly make the HCM soft gripper cheaper, lighter, and faster. The in-plane prestressed HCM fingers work as force amplifiers as well as load-bearing skeletons, enabling them to manipulate both large and small, thick and thin, and rigid and limp objects with speed and strength. Since the HCM fingers mimic the kinematics of the thumb-index finger pinching, the gripper can not only pick up tangible objects like pens and beakers but also can manipulate thin and soft objects like fabrics. The semi-rigid plastic sheets used for the HCMs function as nails to exert force when dealing with limp objects, and additional silicone rubber coatings can work as finger pulps to elevate the friction coefficient, as in Fig. 1A-1E. Figure 1F shows the

*Research supported by the Dept. of Earth and Environmental Engineering, Columbia University.

[1]Zechen Xiong is with the Dept. of Earth and Environment Engineering at Columbia University, New York, NY 10027 USA (phone: 9173023864, e-mail: zx2252@columbia.edu).

[2]Zihan Guo, Li Yuan, Yitong Liu, Yufeng Su, and Hod Lipson are with the Dept. of Mechanical Engineering at Columbia University, New York, NY 10027 USA (e-mail: ys3399@columbia.edu, zg2450@columbia.edu , yl5185@columbia.edu, and hl2891@columbia.edu).

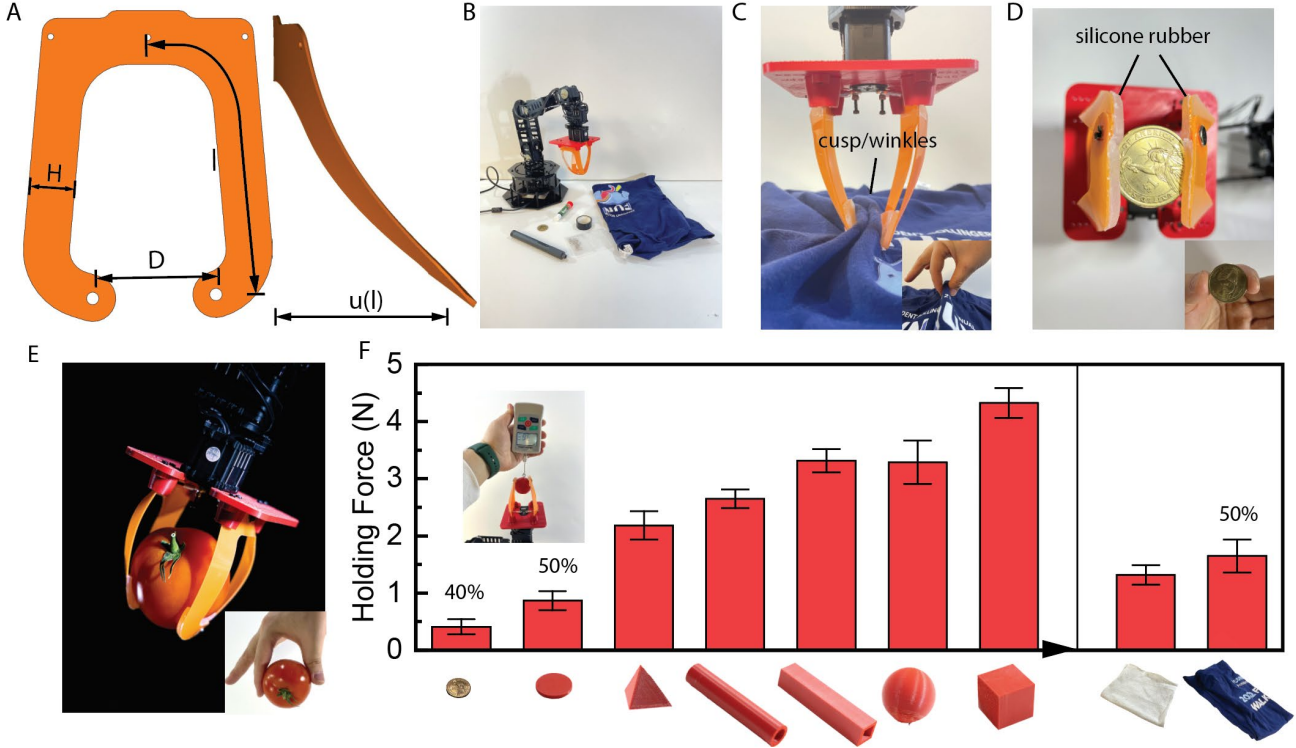

Fig. 1 Bistable hair-clip mechanisms (HCMs) as fingers for a universal soft gripper. (A) The geometry of an HCM finger and its morphology upon assemble. Parameters H is ribbon width, $D$ is the prestressing distance, $l$ is the effective half length of the ribbon, and $u(l)$ is the lateral defection of the HCM mechanism at $z = l$. (B) The HCM fingers installed on a commercial robotic arm for picking up a variety of 2D and 3D objects. The HCM gripper performs thumb-index finger pinching on (C) a t-shirt (front view), (D) a dollar coin (bottom view), and (E) a tomato. The wrinkles or cusp of the t-shirt indicate the different picking-up mechanism for 2D limp objects. Insets show the pinching motion of a human hand for comparison. (F) Holding force (self-weight subtracted) vs. dimension for a wide range of 2D and 3D objects when prestressing level $D = 40$ cm. The left region includes 3D geometric shapes ordered by dimensions and right region shows limp 2D objects including a piece of linen cloth and a t-shirt. The dollor coin has a diameter of 26.5 mm, and can be used for size comparison for the rigid objects. The coin, the thin disc, and the t-shirt have a low picking-up reliability of 40 %, 50 %, and 50 % based on 10 trials, respectively, due to their shapes or self-weight, while others have a success rate of 100%.

holding force, measured with objects' self-weight subtracted, as well as successful rates when the gripper picks up some common objects.

It is estimated that 80% of the total production line time in the garment industry is spent on the handling and manipulation, i.e., the pick-and-place and change of initial shapes of fabrics [18]. To automate the apparel manufacturing industry, a soft gripper that is capable of rapidly manipulating fabrics without damaging them is very attractive. Existent studies demonstrate a series of soft robotic methods that are specified for fabrics. Such methods include pins/needles, vacuum suction, air-jet, gluing, electrostatic [18]–[21], etc. For better efficiency, they sacrifice certain levels of adaptivity and compatibility. To address that and to offer new methodologies, we show in this work that a universal soft gripper with the capability of fabric manipulation is approachable.

The rest of the paper will be organized as follows. We explain the working principle of the HCM, deriving the theoretic solutions to the mechanical properties of the grasping of the HCM fingers in Section II. Then we elaborate on the materials, designs, simulation, fabrication, and measurements for the HCM grippers in Section III. After that, we describe the applications of HCM grippers in fabric manipulation and pneumatic adaptation, and compare them to existing studies in Section IV. Finally, we conclude in Section V.

## II. WORKING PRINCIPLES

### A. Theory of Hair-Clip Mechanism (HCM)

It is proved in our previous work that an HCM can be efficiently used as a skeleton and force amplifier for compliant fish robots and terrestrial crawlers and beat previous records with a three-fold and a five-fold improvement, respectively [16], [22]. The bi-stability of HCM plastic ribbons can increase the angular velocity of a robotic fishtail by about 3 times during snapping [16] and the dome-like structure of the ribbons from their buckled configuration can increase their bending rigidity by about 11 times [22]. All of these features contribute to the performance of our HCM gripper in this paper. And to actuate the HCM fingers, only a point displacement at the center of the ribbon is need, like in Fig. 1A and 1C, simplifying the motion transmission design.

Due to the same theory used, the shape and properties of HCM grippers can be derived similarly as in [16], [22], [23]. Thus, the timescale of the snapping process of the HCM fingers can be estimated as

$$t_* = \frac{(2l)^2}{t\sqrt{E/\rho_s}}, \qquad (1)$$

where $l$ is the effective length of the half ribbon in Fig. 1, $E$ is the material modulus, $t$ is the thickness of the material, and $\rho_s$

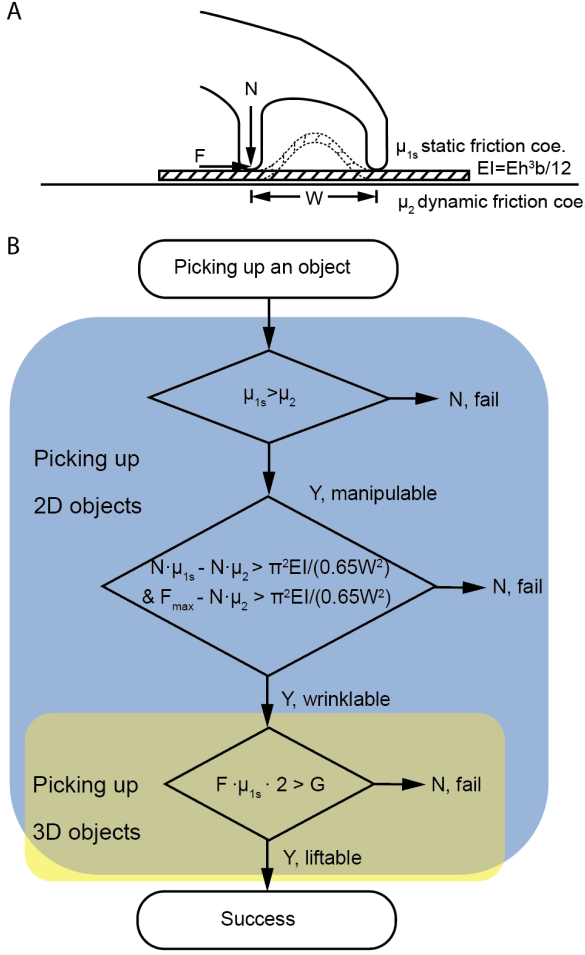

Fig. 2 The difference in criteria for picking up 2D and 3D objects. (A) Force analysis of the thumb-index finger grasping of limp objects. (B) Requirements of picking up 2D and 3D objects. The wrinklable prerequisite requires the maximum friction force N·$\mu_{1s}$ and maximum pinching force $F_{max}$ both to be larger than the Euler's critical load $\pi^2 EI / (0.65W^2)$ of the fabric.

is the density of the material. The opening gap $W$ between the two fingers has a dimension of

$$W = L_f + 2 \cdot u(l) = L_f - 2 \cdot \frac{P_{cr}}{EI_\eta} \int_0^l \int_0^a \varphi(z)(l-z)\,dz\,da, \quad (2)$$

in which $L_f = 48.0$ mm is the installation gap between the fingers, $u(l)$ is the lateral deflection $u(z)$ at $z = l$, and the critical load $P_{cr}$ of the ribbon's lateral-torsional buckling [24] and the lateral rotational angle $\varphi$ are, respectively,

$$P_{cr} = \frac{5.5618}{l^2} \cdot \sqrt{EI_\eta C}, \quad (3)$$

$$\varphi(z) = \sqrt{l-z}\, A_1 J_{1/4}\left(\frac{1}{2}\sqrt{\frac{P_{cr}^2}{EI_\eta C}}(l-z)^2\right), \quad (4)$$

$C = Ght^3/3$ is the torsional rigidity of the ribbon with $G$ the shear modulus and $h$ the width of the ribbon, $A_1$ is a non-zero integration constant derived from energy conservation, $J_{1/4}$ is the Bessel equation of order ¼, and $I_\eta = h^3 t/12$ is the area moment of inertia of the ribbon. The energy released in each grasp of the gripper equals the energy gap between the bi-states of the fingers which is

$$U_{barr} = 6 P_{cr} \cdot D. \quad (5)$$

This energy will dissipate in friction and collision, resulting in high force and high speed.

### B. Grasping limp or fabric-like objects

We propose to use the bi-stable HCMs to mimic the fabric grasping motion of human fingers as in Fig 1C. To the best of the author's knowledge, this is the first universal soft gripper that can deal with objects that are stiff, soft, fabric-like, or tender (e.g., fruits), just like human fingers do. In comparison, jamming grippers can do most rigid objects but none of the thin or fabric-like things [6]. Vacuum and Bernoulli grippers are able to deal with most of the mentioned objects but still have limitations on the shape of the samples. The challenge of manipulating limp objects is due to the disparate criteria between grasping 2D objects and 3D objects, which is illustrated in Fig. 2. Before the "liftable" condition that friction 2 $F \cdot \mu_{1s}$ > object gravity $G$, picking up limp objects has two more preconditions: the friction force of the grasping should be larger than the friction force at the bottom of the fabric; the grasping force should be larger than the bottom friction plus the Euler's critical load $\pi^2 EI / (0.65W^2)$ of the fabric. We term the two preconditions as "manipulable" and "wrinklable" prerequisites. Based on this theory, the properties of our bi-stable soft grippers are suitable for this task: the HCM fingers have a large opening gap $W$ that is usually several times the actuation distance, offering enough space to wrinkle and grasp the cusp of the fabric, and the snap-through buckling can generate a large instant force $F_{max}$ that is unusual in soft grippers, helping overcome friction and resistance. Moreover, bi-stability offers grasping force $F$ without energy input, which is beneficial in long-term manipulation.

### III. FABRICATION AND MEASUREMENT

The HCM gripper is composed of two plastic HCM fingers (McMaster-Carr, 9513K125), a 3D-printed clamping panel (Ultimaker S3, PLA), and a robotic arm (Trossen Robotics, WidowX), as shown in Fig. 1 and Movie S1. An energy source of 12 V is needed for the operation of the actuators and microcontroller (Dynamixel, MX-64, MX-28, AX-12A, ArbotiX Robocontroller). Although multiple materials from paper to steel sheets can be used for HCM designs, we choose semi-rigid plastic material with a modulus of $E = 1730$ MPa [16] to ensure safe interaction and easy fabrication. The HCM fingers are laser cut (Thunder Laser, Nova 51) PETG ribbon with a width of $h = 15$ mm, a thickness of $t = 0.762$ mm, and an effective half-length of $l = 93.7$ mm. To achieve the prestressed configurations, they are pin-locked with press-fit rivets (McMaster-Carr, 97362A), and are coated with silicone rubber sheets (Smooth-On Inc., Ecoflex 10) to increase the friction coefficient $\mu_{1s}$. The dome-like configuration (Fig. 1A) of the HCM fingers offers the rigidity and force needed to grasp the objects. Thus, no additional energy is needed to maintain the closing mode of the gripper.

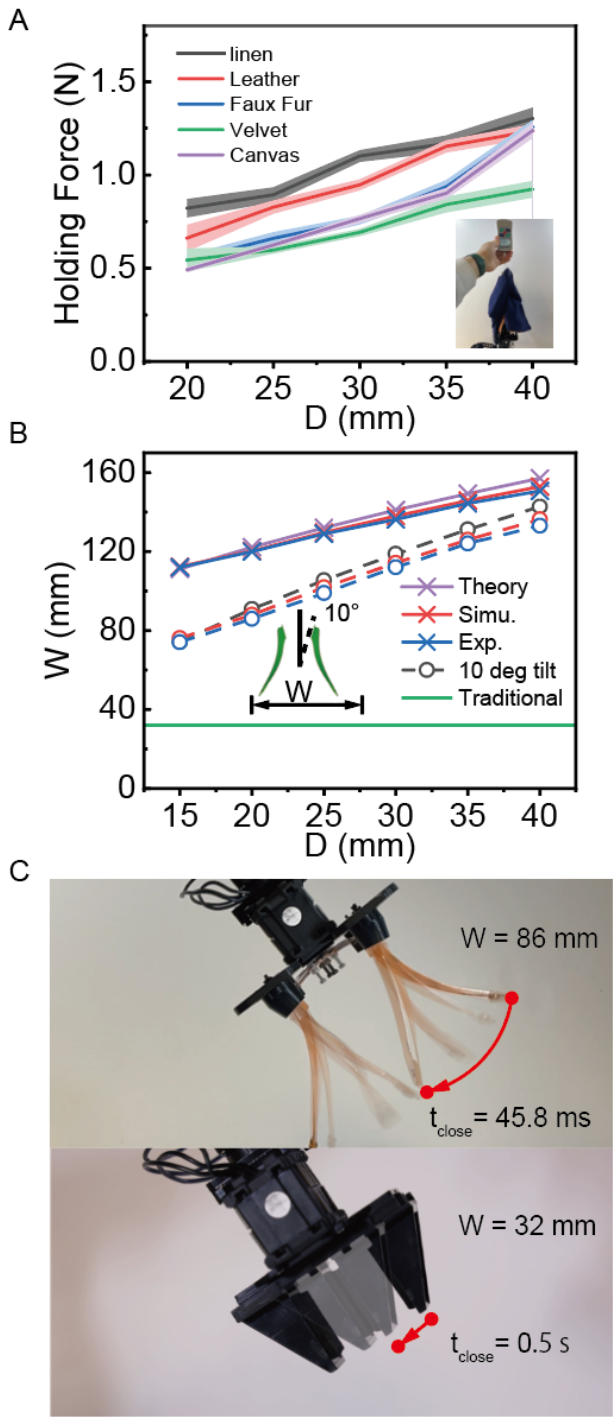

3A, Fig. 1F shows the magnitude of pinching force with respect to the shapes, dimensions, and materials of the objects. Due to the silicone rubber coating (Fig. 1D) and the compliant and curved shape of the HCM fingers, two mechanisms contribute to the holding force: friction and geometric interlocking [6]. For a prestressing level of $D$ = 40 mm, the gripper has grasping forces that are 2.0 – 4.5 N for most rigid objects with the two pinching fingers. Though, the coin and the coin-like disc present challenges to the gripper with their small dimensions in the vertical direction. Only 40 % and 50 % success rates are observed in picking them up based on ten trials, respectively, and low average holding forces of 0.41 N and 0.87 N are obtained (Fig. 1F). The force of picking up a t-shirt (1.65 N) is larger than that of a single layer of fabric (1.32 N), as is shown in Fig. 1F. And due to its self-weight, a t-shirt can only be picked up by the gripper in 50 % of the cases. Despite that, the HCM gripper has a success rate of 100% in ten-trial tests in picking up objects like columns, cubes, pyramids, spheres, and single-layer fabric sheets from the table thanks to friction and posture/geometric interlocking.

A more detailed summary of grasping fabrics and leather samples is shown in Fig. 3A. Same as in Fig. 1F, the fabrics in Fig. 3A are pinched between the fingers and the force meter is lifting a corner of the square fabric sample with a hook. The holding forces range from 0.5 – 1.3 N, dependent on the thickness and texture of the material, and measured from the total lifting force minus the self-weight. Each data point is calculated from 5 trials. In general, the holding forces for 2D objects are smaller than holding geometric shapes like spheres and cubes. This is because only a small cusp or wrinkled part of the fabric is being held (Fig. 2A) and limp materials can easily slip off. For holding heavier fabric sheets, multiple fingers or HCM arrays can be used [25].

### B. Span and Speed

For pinching grippers, the span is important because it determines how large objects the gripper can manipulate. And in the case of fabric grasping, a larger span makes the grasping of limp objects easier since it quadratically decreases the force $F_{max}$ needed to buckle the sheets and also enables the gripper to take hold of a larger cusp or more wrinkles to have a firmer grasp. However, span and speed usually contradict each other for the traditional counterpart (e.g., Trossen Robotics, WidowX) since it generally takes longer to cover a larger distance. Here, we demonstrate that by leveraging bi-stability, our soft gripper can take a grip rapidly, regardless of the large span.

The simulated results of grip span $W$ vs. the prestressing level $D$ are shown in Fig. 3B, together with the theoretical and experimental outcomes. The finite-element (FE) analyses of the configuration (Fig. 1A and Fig. 3B) of the in-plane prestressed HCM are created using ABAQUS/CAE and solved with ABAQUS/Standard. All simulations use about 4600 shell element S4R or S3R and the material is assumed to be linear elastic with Young's modulus ($E$) of 1730 MPa and a Poisson ratio ($v$) of 0.38.

Based on the comparison, the simulative results are quite close to the measurements, giving an error of about 5%. And the theoretical solutions calculated from Eq. (2) are within a 10% error compared to the measurements. The span of our

Fig. 3 The key properties of the HCM grippers. (A) Holding forces of different fabric and leather samples w.r.t the prestressing level $D$. Each point is calculated from five trials. (B) The gripper span $W$ vs. prestressing level $D$ in normal and 10°-tilt situations compared to the span of the traditional counterpart. (C) The comparison of dynamics between the HCM gripper and its traditional counterpart. The HCM gripper with $D$ = 20 cm and a 10° inward tilt of the fingers can snap in 45.8 ± 6.7 ms and cover a span of 86 mm, while the traditional finger covers a span of 32 mm in 0.5 s.

### A. Holding Force

With a digital force gauge (Mxmoonfree, ZMF-50N) and a motion sequence of "picking up from the table (Fig. 1C)" and "lifting to a upward position" shown in Fig. 1D and Fig.

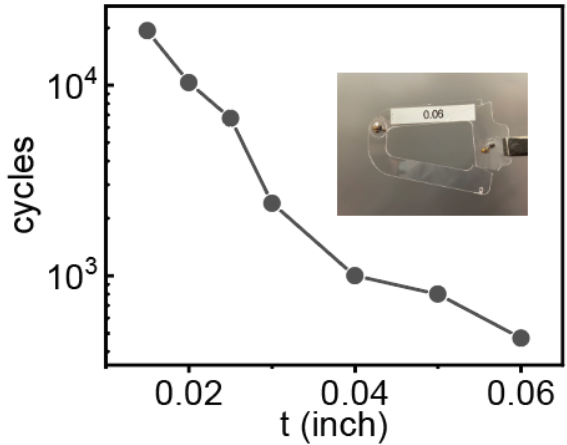

Fig. 4 The fatigue lives of HCM fingers with $D = 20$ mm (a larger $D$ like 40 mm may crack thick HCMs in the prestressing stage) made of PETG plastic sheets with thickness $t = 0.015 – 0.06$ inch ($0.381 – 1.524$ mm). The inset shows the development of crackings at the corners of a plastic HCM finger with $t = 0.06$ inch.

HCM gripper is several times larger than that of the original robotic gripper (Trossen Robotics, WidowX) in Fig. 3B, which is only 32 mm. In some experiments, an inward tilting angle is applied when installing the HCM fingers to increase the holding force, which can decrease the gripper span (Fig. 3B). For example, Fig. 3C presents an HCM gripper with a span of $W = 86$ mm and a tilting angle of 10° that can snap during a short amount of time of $45.8 ± 6.7$ ms, close to the theoretical solutions of 53 ms calculated from Eq. (1), while the closing of the traditional counterpart with a span of $W = 32$ mm usually takes 0.5 s. The HCM is 10.9 times faster than the reference gripper and can vastly increase the grasping efficiency in the garment industry. Movie S1 shows a 2 Hz open-close cycling of the HCM gripper and the major limitation for further increasing the cycling frequency of the gripper is the rotational speed of the motor (59 rev/min, DYNAMIXEL, AX-12A). The trajectory of the HCM fingers is a curved one, in which the HCM fingers first slightly open during the elastic energy storing stage and then suddenly close, similar to the motion of human fingers in a thumb-index finger pinching, while the reference gripper moves straightly with a uniform velocity.

*C. Fatigue*

For compliant mechanisms that incorporate prestressing and bi-stability, the fatigue problem of the material is essential since their motion comes from the bending deformation of flexible parts which leads to stress at those locations [26]. This is even more so when the mechanism needs to be actuated at a high frequency. However, the HCM fingers now use PETG plastic which is not specified for repetitive-loading applications. Some more fatigue-resistant materials like carbon fiber composites [27] can be used for future studies. In Fig. 4, we present the fatigue lives of HCM fingers with prestressing $D = 20$ mm made of PETG plastic sheets of thickness $t = 0.015 – 0.06$ inch ($0.381 – 1.524$ mm). It was found that the PETG plastic sheets can withstand 460 ~ 20000 times cycles depending on the thickness, and that tiny cracks appear mostly at the corner of the HCM fingers. This indicates that PETG plastic may not be a robust material for HCMs; on the other hand, carbon fiber/epoxy plates are found to be a good replacement in our recent studies, which will be described in our future manuscripts.

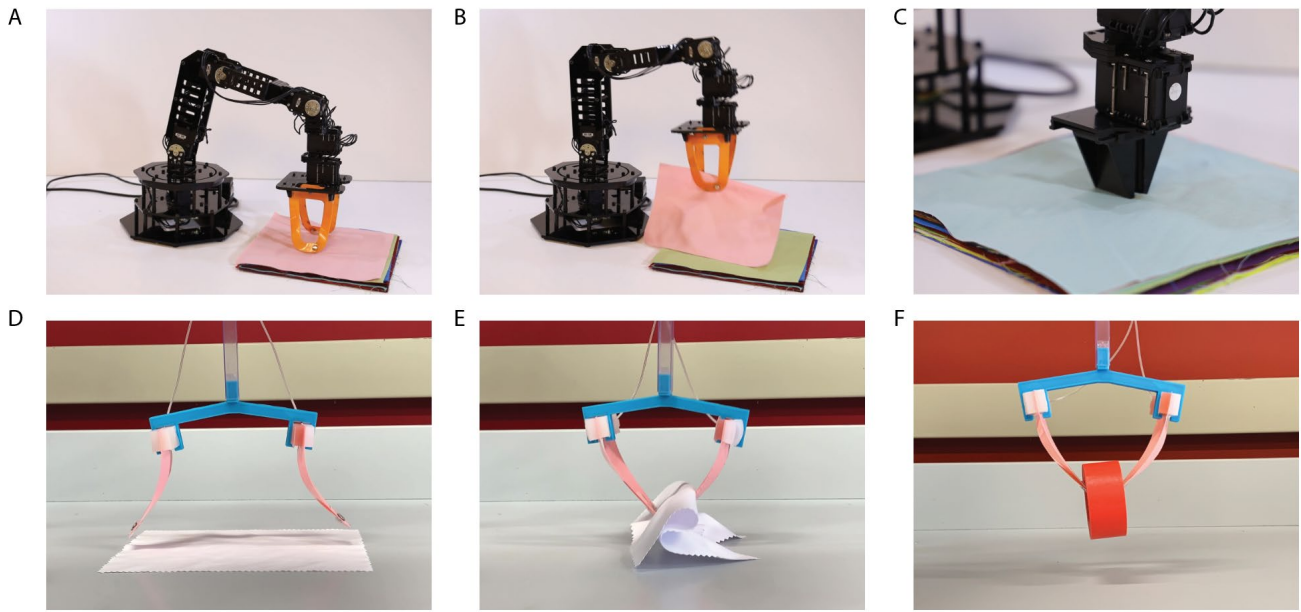

Fig. 5 Manipulating fabric with the motor-driven and the pneumatic HCM grippers. (A) and (B) The process of picking up an initially flat cotton fabric sheet with the motor-driven HCM gripper. The HCM fingers have $D = 20$ mm, $W = 86$ mm, and an inward tilting angle of 10° to increase the grasping force. (C) The failure of the traditional counterpart to manipulate limp materials due to a small span. (D) and (E) The process of pneumatic HCM fingers picking up an initially flat fabric upon an actuation pressure of ~150kPa. The pink plastic has a thickness of $t = 0.381$ mm. (F) Pneumatic HCM gripper is able to pick up a roll of tape (44g), which is three times its self-weight (15g).

## IV. Application

To demonstrate the functionality of the proposed strategy of using pre-stressed plastic HCMs as end-effectors in robotic manipulation, the performance of the motor-driven HCM gripper with a commercial robotic arm and a pneumatic variant of the HCM gripper is presented in this section.

### A. Manipulation of Fabrics

To challenge the gripper's capacity for fabric manipulation and to increase the efficiency of an actual garment production process, we design this test to exhibit the gripper's ability to pick up a single sheet of fabric from a stack of multiple sheets. The picking-up experiment was carried out with $20 \times 20$ cm$^2$ cotton squares. In the demonstration, the gripper moves down to the stack, picks up and removes the top fabric, and places it to the side. Success is defined when the gripper holds only the topmost sheet in the stack and failure otherwise. This process is illustrated in Fig. 5A and 5B and shown in Movie S1. The gripper repetitively picks up the topmost fabric sheet 10 times, and a success rate of ~80% is observed. On the other hand, the traditional counterpart is unable to pick up fabric due to its small span (Fig. 5C and Movie S1).

### B. Pneumatic Gripper

To further demonstrate the compatibility of HCM end-effectors, we build a pneumatic version of it, as in Fig. 5D - 5F and Movie S1. Similar to the motor-driven version, this pneumatic HCM gripper has a minimalistic design that includes two pneumatic-actuated fingers and a 3D-printed panel holding the fingers in position. The pneumatic soft bending units are fabricated with 3D-printed molds and a silicone rubber casting process [16] (Smooth-On Inc., Dragon Skin 20). Upon an actuation pressure of ~150 kPa. It does thumb-index finger pinching and picks up a piece of initially flat soft fabric just like the motor-driven one does. Besides, the pneumatic HCM gripper weighs 15 g and can pick up objects of 44 g (Fig. 5C), which is three times its self-weight. Similarly, the pneumatic variant also only consumes energy when a mode transition is needed, due to its bistable nature.

### C. Comparisons

To explain the role that the HCM gripper plays in the field of soft grippers, a comparison among the proposed HCM soft gripper and a few representative soft grippers [4]–[6], [8]–[10], [12], [19], [20] is given in Table I. It summarizes the grippers' ability to pick up objects with various shapes (universal), to grip rapidly, to manipulate fabric, to have a large span, and to harness bistability. It also mentions whether the gripper uses piercing needles if it can deal with fabrics. Compared to others, the HCM gripper possesses an overall advantage in these mechanical properties, due to its bistable structure and its similarity with human fingers in shape and function. Besides, the HCM fingers are easier to fabricate and assemble than most other grippers. It is also noted that, compared to the commercial reference gripper that shares the same controlling system and the same hardware platform, the HCM modification enables the system to work as a bistable soft gripper with the ability to pick up fabric and significant increases in span and grasping speed.

TABLE I. Comparison among the HCM gripper, reference commercial gripper, and other soft gripper studies

|  | Universal | Fast Grasping | Fabric Picking | Non-Piercing | Large Span | Bi-Stable |
|---|---|---|---|---|---|---|
| **This Work** | Y | 46 ms | Y | Y | 86 mm | Y |
| Ref. | Y | 500 ms | N | -- | 32 mm | N |
| [4] | N | N | N | -- | N | N |
| [5] | Y | N | N | -- | Y | N |
| [6] | Y | N | N | -- | Y | N |
| [8] | N | 112 ms | N | -- | Y | Y |
| [9] | Y | 20 ms | N | -- | Y | Y |
| [10] | Y | 170 ms | N | -- | Y | Y |
| [12] | Y | N | N | -- | Y | Y |
| [19] | N | N | Y | Y | N | N |
| [20] | N | 320 ms | Y | N | N | N |

## V. Conclusion

We present in this work a way of designing and fabricating universal bi-stable grippers that is capable of closing rapidly and picking up limp objects, mimicking the thumb-index finger pinching of human hands. The bi-stable mechanisms we use are termed "hair-clip mechanisms" or HCMs, due to their similarity with and inspiration from a snap hair clip. We depict the design principles using the theoretic solutions obtained from beam buckling theory, fabricate the grippers with PETG plastic sheets, and simulate their deformation using FE software to verify this method. Results show that replacing the end-effectors of the commercial gripper with our HCM fingers can increase the manipulation span by 2.7 times and the grasping speed by 10.9 times. Besides, the HCM fingers can apply gentle grasping forces on a variety of objects including fabrics. A comparison is given between this work and other soft grippers to show the role of this method.

In the future, we are going to expand the use of HCM mechanisms to various applications and solve the present issues such as fatigue and aging in the meanwhile. We hope this study improves the performance of soft robotic manipulators and contributes to the revolution of soft robotics.